\documentclass[10pt,twocolumn,letterpaper]{article}

\usepackage{cvpr}
\usepackage{times}
\usepackage{epsfig}
\usepackage{graphicx}
\usepackage{amsmath}
\usepackage{amssymb}

\usepackage{comment}
\usepackage{color}
\usepackage{url}

\newcommand{\rei}[1]{\textcolor[rgb]{0.0,0,0.0}{#1}}
\newcommand{\yoshi}[1]{\textcolor[rgb]{0.0,0,0.0}{#1}}
\newcommand{\yoshia}[1]{\textcolor[rgb]{0.0,0.0,0.0}{#1}} 
\newcommand{\reia}[1]{\textcolor[rgb]{0.0,0.0,0.0}{#1}} 
\newcommand{\yoshitxt}[1]{\textcolor[rgb]{0.0,0.0,0.0}{#1}} 
\newcommand{\ari}[1]{\textcolor[rgb]{0.0,0.0,0.0}{#1}}
\newcommand{\arm}[1]{\textcolor[rgb]{0.0,0.0,0.0}{#1}}
\newcommand{\narm}[1]{\textcolor[rgb]{0.0,0.0,0.0}{#1}} 
\newcommand{\yoshib}[1]{\textcolor[rgb]{0.0,0.0,0.0}{#1}} 
\newcommand{\shao}[1]{\textcolor[rgb]{0.0,0.0,0.0}{#1}}
\newcommand{\yoshicvpr}[1]{\textcolor[rgb]{0.0,0.0,0.0}{#1}}
\newcommand{\shaocvpr}[1]{\textcolor[rgb]{0.0,0.0,0.0}{#1}}
\newcommand{\arma}[1]{\textcolor[rgb]{0.0,0.0,0.0}{#1}}

\newcommand{\bm}{\boldsymbol}

\newcommand{\mc}{\mathcal}

\def\sysname{CROSR}
\def\datasetnum{five}

\makeatletter
\newcommand{\figcaption}[1]{\def\@captype{figure}\caption{#1}}
\newcommand{\tblcaption}[1]{\def\@captype{table}\caption{#1}}
\makeatother

\usepackage[pagebackref=true,breaklinks=true,letterpaper=true,colorlinks,bookmarks=false]{hyperref}

\cvprfinalcopy 

\def\cvprPaperID{1789} 

\ifcvprfinal\fi
\begin{document}

\title{Classification-Reconstruction Learning for Open-Set Recognition}

\author{Ryota Yoshihashi$^1$ \\　Shaodi You$^{2}$ \\ \vspace{-4mm} \\ $^{1}$The University of Tokyo\\
\and
Wen Shao$^{1}$ \\　Makoto Iida$^{1}$ \\  \vspace{-4mm} \\  \\
\and
Rei Kawakami$^{1}$ \\Takeshi Naemura$^{1}$ \\ \vspace{-4mm} \\$^{2}$Data61-CSIRO
\and 
{\vspace{-10mm} \tt\footnotesize \{yoshi,shao,rei,naemura\}@nae-lab.org, Shaodi.You@data61.csiro.au, iida@ilab.eco.rcast.u-tokyo.ac.jp }
}

\maketitle

\begin{abstract}
Open-set classification is a problem of handling `unknown' classes \arm{that are} not contained in the training dataset, \arma{whereas} traditional classifiers assume that only known classes appear in the test environment. 
Existing open-set classifiers \arm{rely} on deep networks trained \arma{in a supervised manner} on known classes in the training set\arm{;} this causes specialization of learned representations to known classes \arm{and makes it hard} to distinguish unknowns from knowns.
In contrast, we train networks for \yoshicvpr{joint} classification and reconstruction of input data.
This enhances the learned representation \arm{so as to} preserve information useful \arma{for separating} unknowns from knowns, as well as to discriminate classes of knowns. \arm{Our} novel 
{\it Classification-Reconstruction learning for Open-Set Recognition (CROSR)}
utilizes latent representations for reconstruction and enables robust unknown detection
without harming \arm{the} known-class classification accuracy.
\arm{Extensive} experiments reveal that the proposed method outperforms existing deep open-set
classifiers in multiple standard datasets and is robust to
diverse outliers.
The code is available in \url{https://nae-lab.org/~rei/research/crosr/}.
\end{abstract}


\vspace{-4mm}
\section{Introduction}
\vspace{-1mm}
\yoshicvpr{To be deployable \arma{to} real applications, recognition systems} need to be tolerant of unknown things and events that \arma{were} not anticipated during the training phase. 
However, 
most of \arm{the} existing learning methods are \ari{based on}
the closed-world assumption, \ari{that is}, \arma{the} training datasets are assumed to include all classes that appear in the environments where the system will be deployed. \ari{This} assumption can be easily violated
in \arm{real-world} problems, 
\arm{where} covering all possible classes is almost impossible~\cite{McCHay69}. 
Closed-set classifiers are error-prone to samples of unknown classes, and this limits \ari{their} usability~\cite{wilber2013animal,sunderhauf2018limits}.

\begin{figure}[t]
  \begin{center}
    \hspace{-4mm}
    \includegraphics[width=240pt]{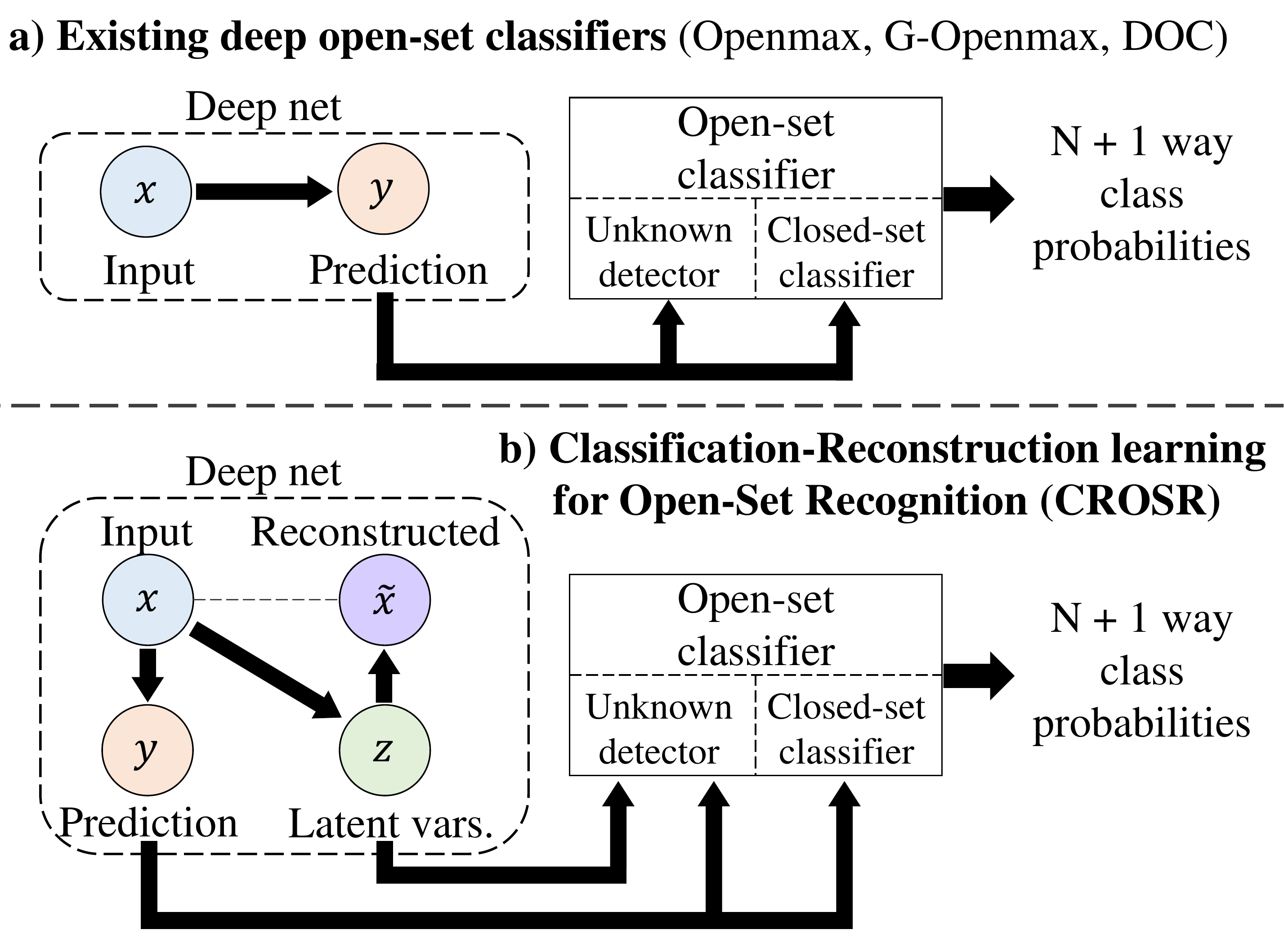}
  \end{center}
  \vspace{-3mm}
  \caption{Overview of existing and our deep open-set classification models. 
  \yoshicvpr{Existing models (a) utilize only their network's final prediction $\bm{y}$ for classification and unknown detection.}
  \yoshicvpr{\arma{In contrast}, in \sysname~(b),} a deep net is trained to provide a prediction $\bm{y}$ and a latent representation for reconstruction $\bm{z}$ within known classes. 
  An open-set classifier (right), \arm{which} consists of an unknown detector and a closed-set classifier, exploits $\bm{y}$ for closed-set classification, and $\bm{y}$ and $\bm{z}$ for unknown detection.}
  \vspace{-5mm}
  \label{fig:overview}
\end{figure}
\arm{In contrast}, 
open-set classifiers~\cite{scheirer2013toward} can detect samples that belong to none of the training classes. 
\reia{Typically, they fit a probability distribution to the training samples in some feature space, and detect outliers as unknowns.} 
\reia{For the features to represent the samples, almost all existing \yoshia{deep} open-set classifiers rely on those acquired via fully supervised learning~\cite{bendale2016towards,ge2017generative,shu2017doc},
\yoshicvpr{as shown in Fig.~\ref{fig:overview} (a).}}
However, they are 
\arm{for emphasizing} the discriminative features of known classes\arm{;} they are not necessarily useful \arm{for representing} unknowns or \arm{separating} unknowns from knowns. 

\reia{In this \arma{study}, our goal is to learn efficient feature representations that are able to classify known classes as well as to detect unknowns as outliers. 
Regarding the representations of outliers that we cannot assume beforehand, it is natural to add unsupervised learning as a regularizer so that the learned representations acquire information that are important in general but may not be useful for classifying given classes.}
Thus, we utilize unsupervised learning of reconstructions in addition to supervised learning of classifications. 
Reconstruction of input samples from low-dimensional latent representations inside the networks is a general way \arma{of} unsupervised learning~\cite{hinton2006reducing}. 
\narm{The \arma{representation learned}} via reconstruction \arm{are} useful \arm{in} several tasks~\cite{zhang2016augmenting}.
Although there are \arm{previous} successful examples of classification-reconstruction learning, such as semi-supervised learning~\cite{rasmus2015semi} and domain adaptation~\cite{ghifary2016deep}, 
\yoshia{this \arm{study} is the first \arm{to} apply deep classification-reconstruction learning to open-set classification.}

\yoshicvpr{Here,
we present a novel open-set classification framework, called Classification-Reconstruction learning for Open-Set Recognition (CROSR).}
\rei{As shown in Fig.~\ref{fig:overview} (b), the open-set classifier consists of two parts: a closed-set classifier and an unknown detector, \yoshicvpr{both of which exploit a deep classification-reconstruction network}.\footnote{We refer to detection of unknowns as {\it unknown detection}, and known-class classification as {\it known classification.}}}
While the known-class classifier exploits \narm{supervisedly learned prediction} $\bm{y}$, 
the unknown detector uses \shao{a reconstructive latent representation}
$\bm{z}$ together with $\bm{y}$.
\yoshi{This allows unknown detectors to exploit a wider pool of features that may not be discriminative for known classes.}
\yoshicvpr{Additionally, in higher-level layers of supervised deep nets, details of input tend to be lost~\cite{zhang2016augmenting,dosovitskiy2016inverting}
, which may not be preferable in unknown detection. CROSR can exploit reconstructive representation $\bm{z}$ to complement the lost information in the prediction $\bm{y}$.}

\yoshicvpr{To provide effective $\bm{y}$ and $\bm{z}$ simultaneously, we further design {\it deep hierarchical reconstruction nets (DHRNets)}.}
\yoshicvpr{The key idea in DHRNets is the bottlenecked lateral connections, which is useful to learn rich representations for classification and compact representations for detection of unknowns jointly.}
DHRNets learn reconstruction of each intermediate layer in classification networks using latent representations, i.e., mapping to low-dimensional spaces, and as a result it acquires hierarchical latent representation.
\yoshicvpr{With the hierarchical bottlenecked representation in DHRNets, the unknown detector in CROSR can exploit multi-level anomaly factors easily thanks to the representations’ compactness.}
\reia{This \yoshicvpr{bottlenecking} is crucial,} 
because \arm{outliers are harder to detect} in higher dimensional feature spaces due to {\it concentration on the sphere}~\cite{zimek2012survey}.
Existing autoencoder variants, \arm{which} are useful for outlier detection by learning compact representations~\cite{zhou2017anomaly,aytekin2018clustering}, cannot afford large-scale classification because  \reia{the bottlenecks in their mainstreams limit the expressive power for classification.}
\yoshicvpr{\sysname~with a DHRNet} 
\yoshicvpr{becomes} more robust to \arm{a} wide variety of unknown samples, 
some of which \ari{are} \arm{very} similar to the known-class samples.
Our experiments in \datasetnum~standard datasets \shao{show} that representations learned via reconstruction serve \narm{to complement those} obtained via classification.

Our contribution is three-fold: First, we discuss \arm{the} usefulness of \yoshia{deep} reconstruction-based representation learning in open-set recognition for the first time\arm{;} all of \arm{the} other deep open-set classifiers are based on discriminative representation learning in known classes.
\yoshi{Second, we develop a novel open-set recognition framework, \sysname, which is based on \shaocvpr{DHRNets} and 
\shao{jointly performs} known classification and unknown detection using them.}
Third, we conducted experiments on open-set classification in \ari{\datasetnum}~standard \arm{image and text} datasets, and \arm{the results} show \arma{that} our method outperforms existing deep open-set classifiers for most combinations of known data and outliers. The code related to this paper is available in \url{https://nae-lab.org/~rei/research/crosr/}.

\vspace{-2mm} \section{Related work} \vspace{-1mm}
\noindent  {\bf Open-set classification}　\hspace{1mm}
Compared \arm{with} closed-set classification, \arm{which has been} investigated \arm{for} decades~\cite{fisher1936use,cortes1995support,freund1997decision}, open-set classification \shao{has} been surprisingly overlooked.
\yoshia{\arm{The few studies on this topic mostly utilized} either linear, kernel, or nearest-neighbor models.
For example, \arma{W}eibull-calibrated SVM~\cite{scheirer2014probability} considers \arma{a} distribution of decision scores for unknown detection. Center-based similarity space models~\cite{fei2016breaking} represent data by their similarity to class centroids \arma{in order to} tighten the distributions of positive data. Extreme value machines~\cite{Rudd_2018_TPAMI} model class-inclusion probabilities using \arma{an} extreme-value-theory-based density function. 
Open-set nearest neighbor \arma{methods}~\cite{junior2017nearest} utilizes \arma{the} distance ratio to the nearest and second nearest classes.}
\yoshia{Among them, sparse-representation-based open-set recognition~\cite{zhang2017sparse} shares \arm{the}
idea of reconstruction-based representation learning with ours.
The difference is in that we consider deep representation learning, 
while~\cite{zhang2017sparse} uses a single-layer linear representation.}
\arm{These models} cannot be applied to large-scale raw data without feature engineering.

The \narm{origin} of deep open-set classifiers \arm{was} \arma{in} 2016~\cite{bendale2016towards}, and 
\arm{few deep open-set classifiers have been reported since then.}
\shao{G-Openmax~\cite{ge2017generative}, a direct extension of Openmax,}
trains networks with synthesized {\it unknown} data by \arm{using} generative models.
However, it \arm{cannot} be applied to natural images other than hand-written characters due to \arma{the} difficulty of generative modeling.
DOC (deep open classifier)~\cite{shu2017doc,shu2018unseen}, which is designed for document classification, enables end-to-end training 
\yoshi{by eliminating outlier detectors outside networks and
using sigmoid activations in the networks for \shao{performing joint} classification and outlier detection.}
Its drawback is that \yoshi{the sigmoids} do not \arm{have the}
{\it compact abating property}~\cite{scheirer2014probability}; 
namely, \rei{they may be activated by an infinitely distant input} from all of the training data, and thus its open space risk is not bounded.

\vspace{1mm} \noindent {\bf Outlier detection}　\hspace{1mm}
\yoshia{Outlier (also called anomaly or novelty) detection can be incorporated \arm{in the concept of open-set-classification} as an unknown detector.
\shaocvpr{However, outlier detectors} are not open-set classifiers by themselves because they have no discriminative power within known classes.} 
Some of the generic methods for anomaly detection are one-class extension of discriminative models such as SVM~\cite{manevitz2001one} or forests~\cite{liu2008isolation}, generative models such as Gaussian mixture models~\cite{roberts1994probabilistic}, and subspace methods~\cite{ringberg2007sensitivity}.
However, \yoshia{most} of the \yoshia{recent} anomaly-detection literature \arm{focuses on} incorporating domain knowledge 
\arma{specific to the task at hand}, such as cues from videos~\cite{xu2015learning,hinami2017joint}, and they cannot be \arm{used} to build a generic-purpose open-set classifiers.

Deep nets \arm{have also been examined} for outlier detection.
The deep approaches mainly use autoencoders \arma{trained in an unsupervised manner}~\cite{zhou2017anomaly},
in combination with GMM~\cite{zong2018deep}, clustering~\cite{aytekin2018clustering},
or one-class learning~\cite{perera2018learning}.
\yoshia{
\shao{Generative adversarial nets~\cite{goodfellow2014generative} \arma{can be used} for outlier detection~\cite{schlegl2017unsupervised} by using their reconstruction errors and discriminators' decisions.}
This usage is different from ours \shaocvpr{that} utilizes latent representations.}
However, in outlier detection, deep nets are not always the absolute winners unlike in supervised learning, because
nets need to be trained \arm{in an unsupervised manner and are less effective because of that}.

\yoshi{Some studies \shao{use} \arma{networks trained in a supervised manner} to detect anomalies \narm{that are not} from the distributions of training data~\cite{hendrycks2016baseline,liang2017enhancing}.}
However, \yoshi{their methods} cannot be simply extended to open-set classifiers because
\shaocvpr{they use input preprocessing,} 
\arma{for example,} 
adversarial perturbation~\cite{goodfellow2015explaining}, \shaocvpr{and this operation} may \arma{degrade} known-class classification.

\vspace{1mm} \noindent {\bf Semi-supervised learning}　\hspace{1mm}
In semi-supervised learning settings including domain adaptation, reconstruction is useful as a data-dependent regularizer~\cite{rasmus2015semi,maaloe2016auxiliary}.
\yoshicvpr{Among them, ladder nets~\cite{rasmus2015semi} are partly similar to ours in terms of using lateral connections, except that ladder nets do not have the bottleneck structure.}
\arm{Our work aims at demonstrating} that the reconstructive regularizers are also useful in open-set classification. 
\yoshia{However, the usage of the regularizers is
largely different; \sysname~uses them to prevent the representations from overly specializing to known classes, while semi-supervised learners use them to incorporate unlabeled data \arma{in} their training objectives.}
Furthermore, in semi-supervised learning settings reconstruction errors
are computed on {\it unlabeled} data as well as labeled training data. In open-set settings, it is \arm{impossible} to compute reconstruction errors on any {\it unknown} data\arm{;} we only use labeled (known) training data.

\vspace{-1mm} \section{Preliminaries} \vspace{-2mm}
\yoshib{Before introducing \sysname, we briefly review Openmax~\cite{bendale2016towards}, the existing  
deep open-set classifier. We also introduce the \arma{terminology and notation.}}

Openmax is an extension of Softmax.
Given a set of known classes $\mc{K} = \{C_1, C_2, ..., C_N\}$ and an input data point $\bm{x}$, 
Softmax is \arm{defined} as \shao{following}: 
\begin{eqnarray}
\bm{y} &=& \bm{f}(\bm{x}), \label{eq:closedcls}\\ \nonumber
p(C_i | \bm{x}, \bm{x} \in \mc{K}) &=& \textrm{Softmax}_i( \bm{y}) =
\frac{\exp(x_i)}{\sum_j^N \exp(x_j)},
\end{eqnarray}
where $\bm{f}$ denotes the network as a function and $\bm{y}$ denotes the representation of its final hidden layer, whose \rei{dimensionality is equal to the number of the known classes.}
\narm{
\shao{To be consistent with}~\cite{bendale2016towards}, we refer to it as the activation vector (AV).}
\yoshi{Softmax is designed for closed-set settings where $\bm{x} \in \mc{K}$,
and in open-set settings, we need to consider $\bm{x} \not\in \mc{K}$.}
This is achieved by calibrating the AV by the inclusion probabilities of each class: 
\begin{eqnarray}
\textrm{Openmax}_i(\bm{x}) \hspace{-0mm} &=& \hspace{-0mm} \textrm{Softmax}_i(\bm{\hat{y}}), \\
\bm{\hat{y}}_i &=& \left\{\hspace{-2mm}
	\begin{array}{ll}
    	\bm{y}_i \bm{w}_i & (i \leq N) \\
        \sum_{i=1}^N  \bm{y}_i(1 - \bm{w}_i) & (i = N + 1),
    \end{array}　\right.  \nonumber
\label{eq:Openmax}
\end{eqnarray}
where $w_i$ represents the belief \arm{that} $\bm{x}$ belongs to the known class $C_i$. Here, $\bm{\hat{y}}$, the {\it calibrated activation vector} prevents Openmax from \arm{giving} high confidences to outliers \yoshib{that \shao{give} small $\bm{w}$}, \arm{i.e.,} the unknown samples that do not belong to $C_i$. Formally, the class $C_{N+1}$ represents the {\it unknown} class. 
Usage of $p(\bm{x} \in C_i)$ can be understood as a proxy for $p(\bm{x} \in \mc{K}) $, \arm{which} is harder to model due to inter-class variances.

\arm{\yoshib{For modeling class-belongingness $p(\bm{x} \in \mc{K}) $, we need a distance function $d(\cdot, \cdot)$ and its distribution. \shao{The distance measures the affinity of a data point to each class.}}
Statistical extreme-value theory suggests that the Weibull family of distribution\arma{s} is suitable~\cite{Rudd_2018_TPAMI} for \arma{this} purpose.}
Assuming that $d$ of \arma{the} \yoshi{inliers} follows a Weibull distribution,
class-belongingness \arm{can be expressed} \yoshi{using the cumulative density function}\arma{,}
\begin{eqnarray} \label{eqn:openmax}
p(\bm{x} \in C_i) \hspace{-2mm} &=& \hspace{-2mm} 1- R_{\alpha}(i) \cdot \textrm{WeibullCDF} (d(\bm{x}, C_i); \rho_i) \label{eqn:weibull} \nonumber \\ 
\hspace{-2mm} &=& \hspace{-2mm} 1 - R_{\alpha}(i) \exp\left(-\left(\frac{d(\bm{x}, C_i)}{\eta_i}\right)^{m_i} \right). 
\end{eqnarray}
Here, $\rho_i = (m_i, \eta_i)$ are parameters of the distribution that are derived from the training data of the class $C_i$.
$R_{\alpha}(i) = \max \left(0, \frac{\alpha - \textrm{rank} (i)}{\alpha}\right) $ is a heuristic calibrator \yoshia{that makes \arma{a} larger discount in more confident classes},
\arm{and is} defined by a hyperparameter $\alpha$. $\textrm{rank} (i)$ \shao{is}
the index in the AV sorted in descending order.

\arm{As a class-belongingness measure, we used \arma{the} $\ell^2$ distance of AVs from \arma{the} class means, similarly to nearest non-outlier classification~\cite{bendale2015towards}:}
\begin{eqnarray}
d(\bm{x}, C_i) = \left| \bm{y} - \bm{\mu}_i \right|_2. \label{eqn:l2}
\end{eqnarray}
\yoshib{This gives a strong simplification assuming that $p({\bf x} \in C_i)$ depends only on the $\bm{y}$.}

\section{CROSR: Classification-reconstruction learning for open-set recognition}\vspace{-2mm}
Our design of {\it \sysname} is based on observations \arm{about} Openmax's formulation:
AVs are not necessarily the best representations for \arm{modeling} the class-belongingness \hbox{$p(\bm{x} \in C_i)$}.
\yoshi{\arm{Although} AVs in supervised networks are optimized to give correct $p(C_i | \bm{x})$,}
they are not encouraged to encode information about $\bm{x}$,
and \arm{it is not sufficient to test whether $\bm{x}$ itself is probable in $C_i$.}
We \arm{alleviate this problem} by exploiting reconstructive latent representations, \yoshib{which
encode more about $\bm{x}$}.


\subsection{Open-set classification with latent representations}\vspace{-1mm}
To \arm{enable the use of} latent \yoshib{representations} for reconstruction in the unknown detector, we \yoshia{extend} the Openmax classifier (Eqns.~\ref{eq:closedcls} --~\ref{eqn:l2}) as \arma{follows}.
\shao{We replace Eqn.~\ref{eq:closedcls} for applying the main-body network $\bm{f}$ to both known classification and reconstruction\arm{:}}
\begin{eqnarray}
(\bm{y}, \bm{z}) &=& \bm{f}(\bm{x}), \nonumber \\ 
p(C_i | \bm{x}, \bm{x} \in \mc{K}) &=& \textrm{Softmax}_i(\bm{y}), \\
\bm{\tilde x} &=& \bm{g}(\bm{z}). \nonumber
\end{eqnarray}
Here we \arm{have} introduced $\bm{g}$, a decoder network only used in training
to make \arm{the} latent representation $\bm{z}$ meaningful via reconstruction. $\bm{\tilde x}$ is the reconstruction of $\bm{x}$ using $\bm{z}$. \arma{These equations correspond} to the left part of Fig.~\ref{fig:overview} (b).

\yoshi{The network's prediction $\bm{y}$ and latent representation $\bm{z}$
are jointly used in \arm{the} class-belongingness modeling. 
Instead of Eqn.~\ref{eqn:l2}, \sysname~considers \arma{the} joint distributions 
of $\bm{y}$ and $\bm{z}$ \arma{to be} a hypersphere per class:}
\begin{eqnarray}
d(\bm{x}, C_i) = \left| [\bm{y} , \bm{z}] - \bm{\mu}_i \right|_2 \label{eqn:oneclass}.
\end{eqnarray}
Here, $[\bm{y}, \bm{z}]$ denotes concatenation of the vectors of $\bm{y}$ and $\bm{z}$, and \shaocvpr{$\bm{\mu}_i$} denotes their mean within class $C_i$.


\begin{figure*}[t]
  \begin{center}
    \includegraphics[width=450pt]{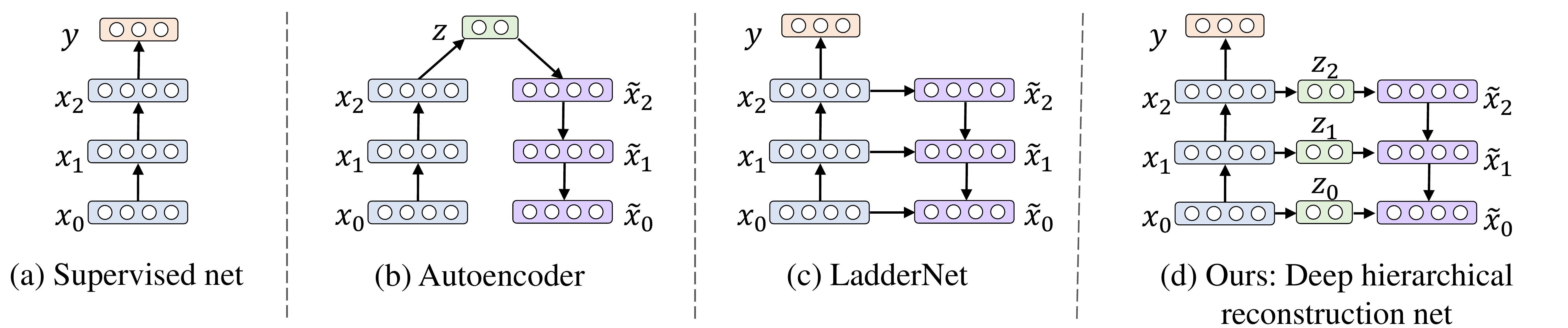}
  \end{center}
  \vspace{-3mm}
  \caption{Conceptual \arm{illustrations} of (a--c) existing models and (d) our model.}
  \vspace{-3mm}
  \label{fig:models}
\end{figure*}
\begin{figure}[t]
  \begin{center}
    \includegraphics[width=245pt]{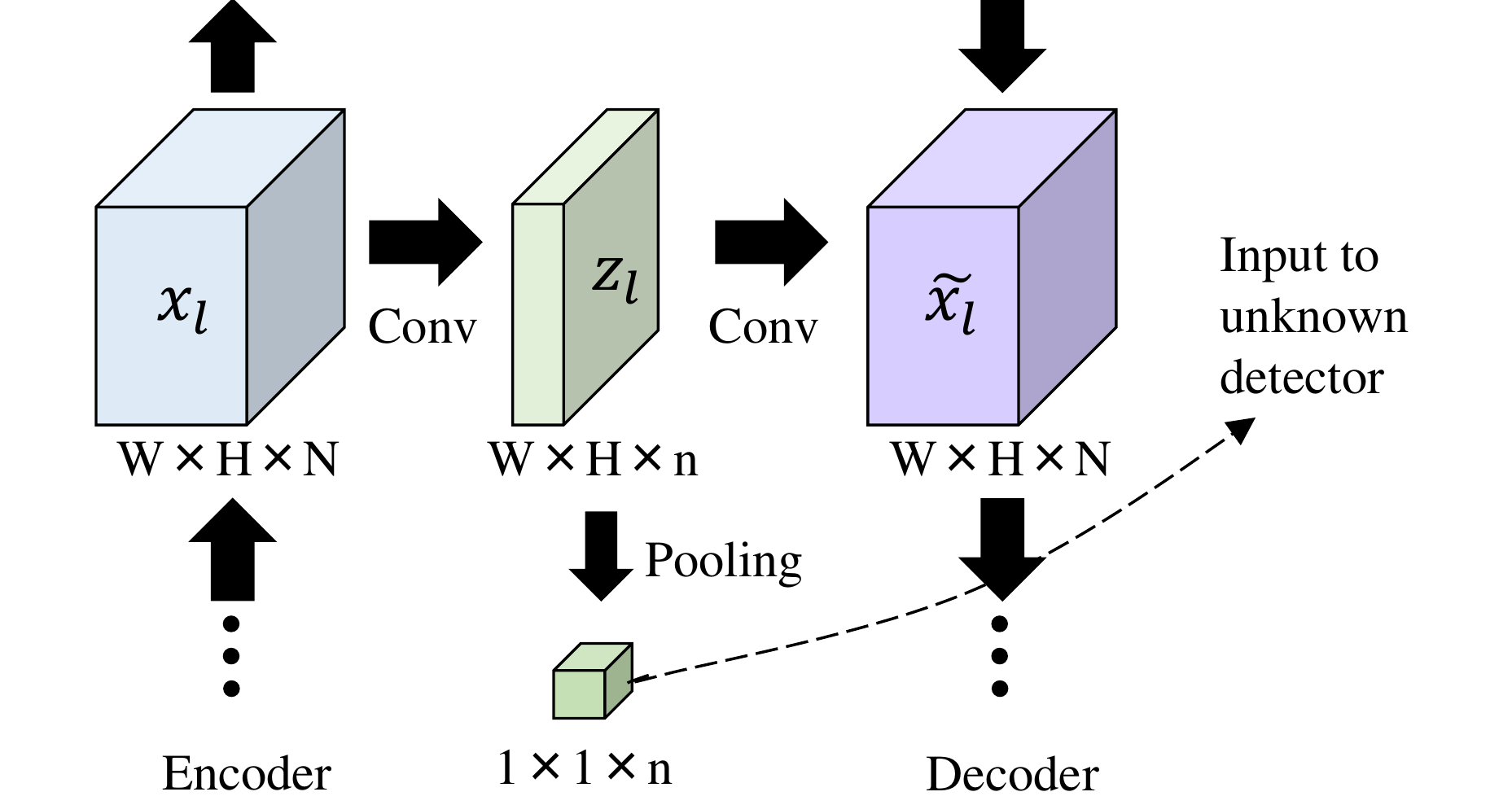}
  \end{center}
  \vspace{-3mm}
  \caption{Implementation of the deep hierarchical reconstruction net with 
  convolutional layers.}
  \vspace{-5mm}
  \label{fig:conv}
\end{figure}

\subsection{Deep Hierarchical Reconstruction Nets}\vspace{-1mm}
After designing the open-set classification framework, 
we \arm{must} specify the function form, \arm{i.e.,} the network architecture for $\bm{f}$.
\yoshib{The network used in \sysname~needs to effectively provide \arma{a} prediction $\bm{y}$ and latent representation $\bm{z}$.}
Our design of \shaocvpr{deep hierarchical reconstruction nets (DHRNets)} simultaneously \arm{maintains} the accuracy of 
$\bm{y}$ in known classification and provides \arma{a} compact $\bm{z}$.

\yoshia{For a conceptual explanation,} \shaocvpr{DHRNet} extracts the latent representations from
each stage of middle-level layers in the classification network.
Specifically, \arm{it} extracts a series of latent representations 
$\bm{z}_1, \bm{z}_2, \bm{z}_3, ..., \bm{z}_L$ from multi-stage features 
$\bm{x}_1, \bm{x}_2, \bm{x}_3, ..., \bm{x}_L$.
\yoshi{We refer to these latent representations as {\it bottlenecks}}.
The advantage of this architecture is that it can detect outlying factors \arm{that are} hidden in the input data but vanish in the middle of \arm{the} inference chains.
Since we cannot presume \arm{a} stage where the outlying factors are most obvious,
we construct the input vector for the unknown detector $\bm{z}$ by simply concatenating
$\bm{z}_l$ from the layers.
Here, $\bm{z}_1, \bm{z}_2, \bm{z}_3, ..., \bm{z}_L$ can be interpreted as decomposed factors to generate $\bm{x}$.
\yoshia{\arm{To draw} an analogy, unknown detection using
decomposed latent representations is similar
to overhauling~\cite{mobley2008maintenance} mechanical products, \narm{where one}
disassembles $\bm{x}$ into parts $\bm{z}_1, \bm{z}_2, \bm{z}_3, ..., \bm{z}_L$, investigates the parts for \arma{anomalies}, 
and reassembles them into $\bm{\tilde x}$.}

Figure~\ref{fig:models} \arm{compares the}
existing architectures and \shaocvpr{DHRNet}.
Most of \arm{the} closed-set classifiers and Openmax rely on supervised classification-only models (a)
that do not have useful factors for outlier detection other than $\bm{y}$,
because $\bm{x}_l$ usually has high dimensionality for known-class classification.
Employing autoencoders (b) is a straightforward \arm{way to introduce
latent representations} for reconstruction, but there is a problem in
using them for open-set classification. 
Deep autoencoders gradually reduce \arm{the} dimensionality of the intermediate layers 
$\bm{x}_1, \bm{x}_2, \bm{x}_3, ..., $ for effective information compression. This is \arm{not good for} large-scale closed-set classification, \arm{which} needs a fairly large number of neurons
in all layers to learn a rich feature hierarchy.
LadderNet (c) can be \arm{regarded} as a variant of an autoencoder, because it performs reconstruction.
However, the difference \arm{lies in the} {\it lateral connections}, 
\yoshicvpr{through which part of $\bm{x}_l$ flows 
to the reconstruction stream without further compression}.
\arm{Their} role is \arma{in a} detail-abstract decomposition~\cite{valpola2015neural}\arm{;} that is, LadderNet encodes
abstract information in the main stream and details in 
the lateral paths. 
\yoshia{While this is preferable \arm{for} open-set classification because the outlying factors of unknowns
may be in \arm{the} details as well as in \arm{the} abstracts, LadderNet itself does not provide
compact latent variables.}
\yoshia{\shaocvpr{DHRNet} (d) further enhances
the decomposed information's effectiveness for unknown detection by compressing the lateral streams in
compact representations $\bm{z_1}, \bm{z_2}, ..., \bm{z_L}$.}

\yoshi{\yoshib{In detail}, the} $l$-th layer of \shaocvpr{DHRNet} is \arm{expressed} as
\begin{eqnarray}\label{eqn:encdec}
\bm{x}_{l + 1} &=& \bm{f}_l (\bm{x}_{l}), \nonumber \\
\bm{z}_{l} &=& \bm{h}_l(\bm{x}_{l}), \\
\bm{\tilde x}_{l} &=& \bm{g}_l(\bm{\tilde x}_{l + 1} + \bm{\tilde h}_l(\bm{z}_{l})). \nonumber 
\end{eqnarray}
Here, $\bm{f}_l$ denotes a block of a feature transformation in the network, i.e.,
a series of convolutional layers between downsampling layers in a plain CNN or a densely-connected block in DenseNet~\cite{huang2017densely}.
$\bm{h}_l$ \arm{denotes} an operation of non-linear dimensionality reduction, 
which consists of a ReLU \arma{and a} convolution layer, 
while $\bm{\tilde h}_l$ means a reprojection to the original dimensionality of $\bm{x}_l$.
The pair of $\bm{h}_l$ and $\bm{\tilde h}_l$ is similar to an autoencoder.
$\bm{g}_l$ is a combinator of the top-down information $\bm{\tilde x}_{l+1}$ 
and lateral information $\bm{\tilde h}_l(\bm{z}_{l})$.
While the function forms for $\bm{g}_l$ \shao{are} investigated by~\cite{pezeshki2016deconstructing},
we \shao{choose} to use an element-wise sum and \narm{subsequent} convolutional and ReLU layers
as the simplest form among the possible variants.
When inputting $\bm{z}_{l}$ to the unknown detectors, the spatial axes are reduced by 
global max pooling to form a one-dimensional vector. 
This performs slightly better than vectorization \yoshi{by using average pooling or flattening.} 
\yoshib{Figure \ref{fig:conv} illustrates these \arma{operations},
and the stack of operations gives \arm{the} overall network shown in Fig. \ref{fig:models} (d).}

\vspace{1mm} \noindent  {\bf Training}　\hspace{1mm}
\yoshicvpr{We minimize the sum of classification errors and reconstruction errors \shaocvpr{in} training data from known classes.
To measure the classification error, we use softmax cross \narm{entropy} of $\bm{y}$ and the ground-truth labels.
\yoshitxt{To measure the reconstruction error of $\bm{x}$ and $\bm{\tilde x}$}, we use \arma{the} $\ell^2$ distance in \arma{the} images and \arma{the} cross entropy of one-hot word representations in \arma{the} texts.}
Note that we cannot use the data of the {\it unknown} classes in training and 
the reconstruction loss is computed only with known samples.
The whole network is differentiable and trainable using gradient-based methods.
\yoshi{After the network is trained and its weights fixed, we compute Weibull distributions for unknown detection.}

\vspace{1mm} \noindent  {\bf Implementation}　\hspace{1mm}
There are some more minor differences \arm{between} our implementation and the ladder nets in
\cite{rasmus2015semi}. 
\yoshia{First, we use dropout in intermediate layers instead of noise addition, 
because it results in slightly better closed-set accuracy.}
Second, we \narm{do} not penalize reconstruction errors of intermediate layers.
\yoshicvpr{This enables \arma{us to avoid the} separate computation of 'noisy' and 'clean' layers that was originally needed for intermediate-layer reconstruction.}
\yoshi{We simply refer to our network without bottlenecks; in other words where $\bm{h}_l$ and $\bm{h}'_l$ are identity transformations, as LadderNet.}
\arm{For} \arma{the} experiments, we implement LadderNet and \shaocvpr{DHRNet} with various backbone architectures.

\vspace{-2mm}
\section{Experiments}\vspace{-2mm}
We experimented \arm{with} \sysname~and other methods \arm{on} \yoshitxt{\datasetnum} standard datasets: 
\yoshitxt{MNIST, CIFAR-10, SVHN, TinyImageNet, and DBpedia}.
\yoshib{\arma{These} datasets are for closed-set classification}, \arma{and} we extended \arma{them} in two \arma{ways}: 1) class separation and 2) outlier addition.
In class-separation setting, 
\yoshib{we selected some classes randomly \arma{in order to} use them as knowns\arma{. We used the remainder} as unknowns.}
\yoshicvpr{\arma{In this setting, which has been used in the} open-set literature~\cite{shu2017doc,neal2018open},}
\arm{unknown samples come from the same domain as that of knowns.}
Outlier addition is a protocol introduced for out-of-distribution detection~\cite{hendrycks2016baseline}\arm{; 
the networks are trained on} the full training data\arm{, but in the test phase,}
outliers from another dataset are added \arma{to} the test set as {\it unknowns}.
The merit \arma{of doing so} is that we can test \arma{the} robustness of the classifiers \arm{against a larger diversity of data than in the original datasets.} 
The class labels of \arma{the} {\it unknowns} were not used \arma{in any case} and
they all were treated as a single {\it unknown} class.

\begin{table}[t]
	\caption{Closed-set test accuracy of used networks. Despite adding reconstruction terms \arm{to the} training objectives for LadderNet and \shaocvpr{DHRNet}, there was no significant degradation \yoshib{ \arma{in} accuracy in known classification}.}
  \vspace{-2mm}
   \footnotesize
  \begin{center}
  \begin{tabular}{|c|c|c|c|c|c|} \hline
    &                 & MNIST & C-10 & SVHN \\ \hline \hline
  	Plain CNN & Supervised only & 0.991 & 0.934 &  0.943   \\
               & LadderNet         & 0.993 & 0.928 &  --    \\
               & DHRNet (ours)     & 0.992 & 0.930  &  0.945   \\ \hline
    DenseNet & Supervised only  & --    & 0.944 &  --     \\
               & DHRNet (ours)     & --    & 0.940  &  --      \\ \hline
  \end{tabular}\vspace{-8mm}
  \end{center}
  \label{tab:closedacc}
\end{table}

\begin{figure*}[t]
  \begin{tabular}{cc}
	\begin{minipage}[t]{135pt}
    	\begin{center}
          \includegraphics[width=135pt]{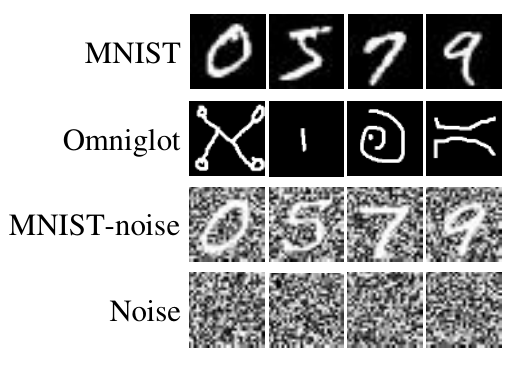}
        \end{center}
        \vspace{-4mm}
        \caption{Sample images from MNIST and outlier sets.}
        \label{fig:samples}
	\end{minipage} \hspace{4mm} \begin{minipage}[t]{325pt}
          \vspace{-32mm}
          \def\@captype{table}
          \tblcaption{Open-set classification results in MNIST with various outliers added \arm{to} the test set as {\it unknowns}. \arm{We report macro-averaged F1-scores in eleven classes 
          ({\it 0--9} and {\it unknown}).} \arma{A larger score} is better.}\vspace{-5mm}
        \footnotesize
        \begin{center}
        \begin{tabular}{|c|c|c||c|c|c|} \hline
           Backbone network & Training method & UNK detector & Omniglot & MNIST-noise & Noise \\ \hline \hline
           Plain CNN & Supervised only  & Softmax           & 0.592 & 0.641 & 0.826 \\
                     &                  & Openmax           & 0.680 & 0.720 & {\bf 0.890} \\ 
                     & LadderNet           & Softmax           & 0.588 & 0.772 & 0.828\\ 
                     &                  & Openmax           & {0.764} & 0.821  & 0.826\\ 
                     & DHRNet (ours) & Softmax            & 0.595 & 0.801  & 0.829\\
                     &                  & Openmax            & 0.780 & 0.816  & 0.826\\ 
                     &                  & \sysname~(ours)     & {\bf 0.793} & {\bf 0.827} & 0.826\\ 
           \hline
        \end{tabular}\vspace{-2mm}
        \end{center}
        \label{tab:resmnist}
      \end{minipage}
  \end{tabular}
\end{figure*}

\begin{table*}[t]
    \vspace{-1mm}
	\caption{Open-set classification results in CIFAR-10. \arma{A larger score} is better.}
  \footnotesize
  \begin{center}
  \begin{tabular}{|c|c|c||c|c|c|c|} \hline
  	 Backbone network & Training method & UNK detector & ImageNet-crop & ImageNet-resize & LSUN-crop & LSUN-resize \\ \hline \hline
     Plain CNN & \multicolumn{2}{c||}{Counterfactual\,\cite{neal2018open}}  & 0.636 & 0.635 & 0.650 & 0.648 \\ \hline
    Plain CNN      & Supervised only  & Softmax    & 0.639 & 0.653 & 0.642 & 0.647 \\
               &                  & Openmax    & 0.660 & 0.684 & 0.657 & 0.668 \\ 
               & LadderNet           & Softmax    & 0.640 & 0.646 & 0.644 & 0.647 \\ 
               &                  & Openmax    & 0.653 & 0.670 & 0.652 & 0.659 \\
               &                  & CROSR      & 0.621 & 0.631 & 0.629 & 0.630 \\
               & DHRNet (ours) & Softmax    & 0.645 & 0.649 & 0.650 & 0.649 \\ \
               &                  & Openmax    & 0.655 & 0.675 & 0.656 & 0.664 \\ 
               &                  & \sysname~(ours) & {\bf 0.721} & {\bf 0.735} & {\bf 0.720} & {\bf 0.749} \\ \hline
     DenseNet  & Supervised only  & Softmax & 0.693 & 0.685 & 0.697 & 0.722 \\
               &                  & Openmax & 0.696 & 0.688 & 0.700 & 0.726 \\ 
               & DHRNet (ours) & Softmax & 0.691  & 0.726 & 0.688 & 0.700 \\ \
               &                  & Openmax & 0.729  & 0.760 & 0.712 & 0.728 \\ 
               &                  & \sysname~(ours) & {\bf 0.733} & {\bf 0.763} & {\bf 0.714} & {\bf 0.731} \\
     \hline
  \end{tabular} \vspace{-6mm}
  \end{center}
  \label{tab:resc10}
\end{table*}

\begin{figure}[t]
  \begin{center}
    \hspace{-4mm}
    \includegraphics[width=225pt]{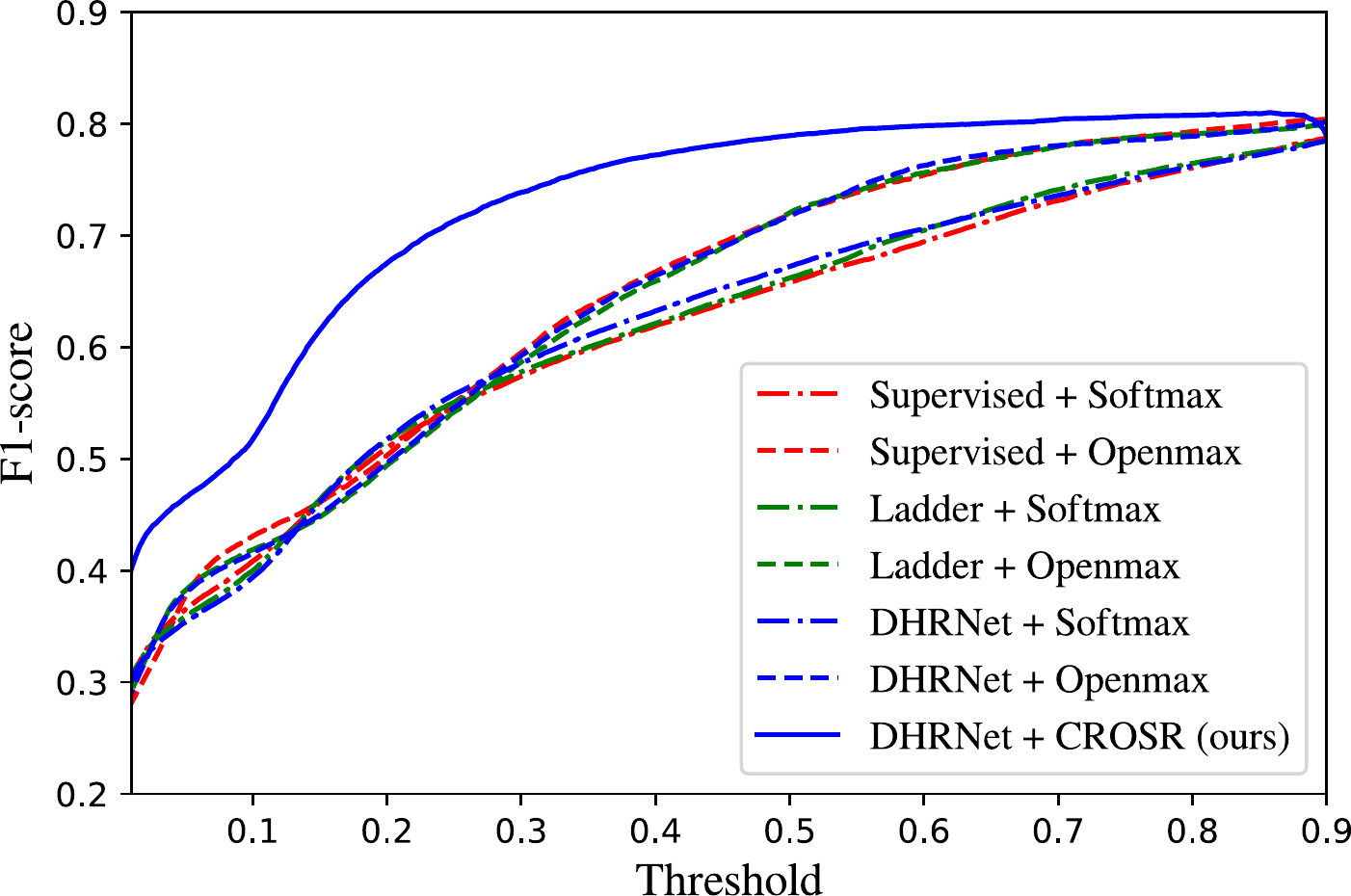}
  \end{center}
  \vspace{-4mm}
  \caption{Relationship between the rejection threshold and F1-score.
  These plots are from test results for CIFAR-10 and ImageNet-crop using VGGNets.}
  \vspace{-2mm}
  \label{fig:f1_th}
\end{figure}

\begin{figure*}[t]
  \begin{center}
    \includegraphics[width=470pt]{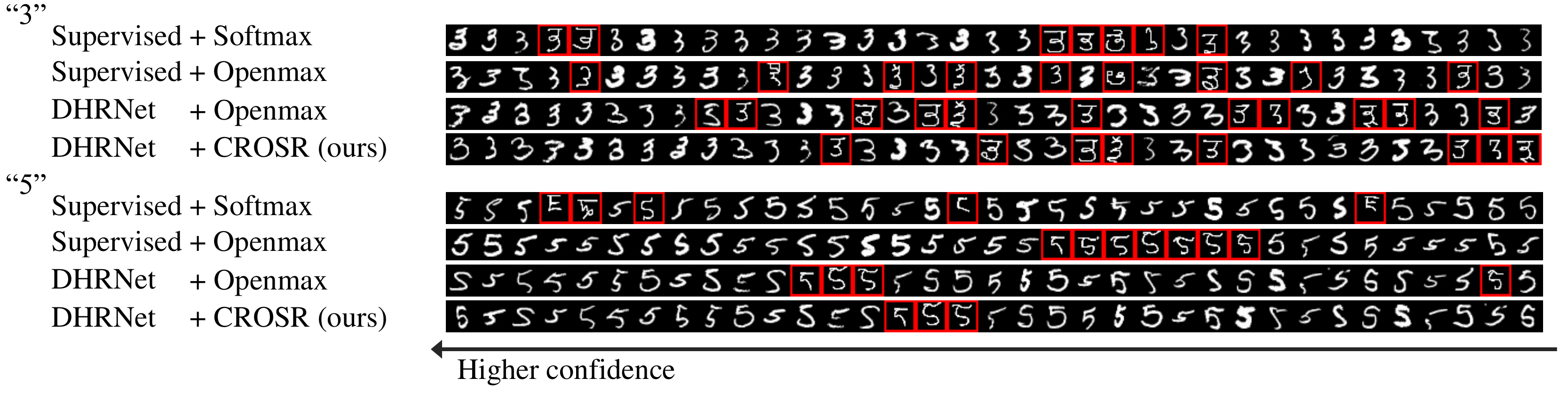}
    \includegraphics[width=470pt]{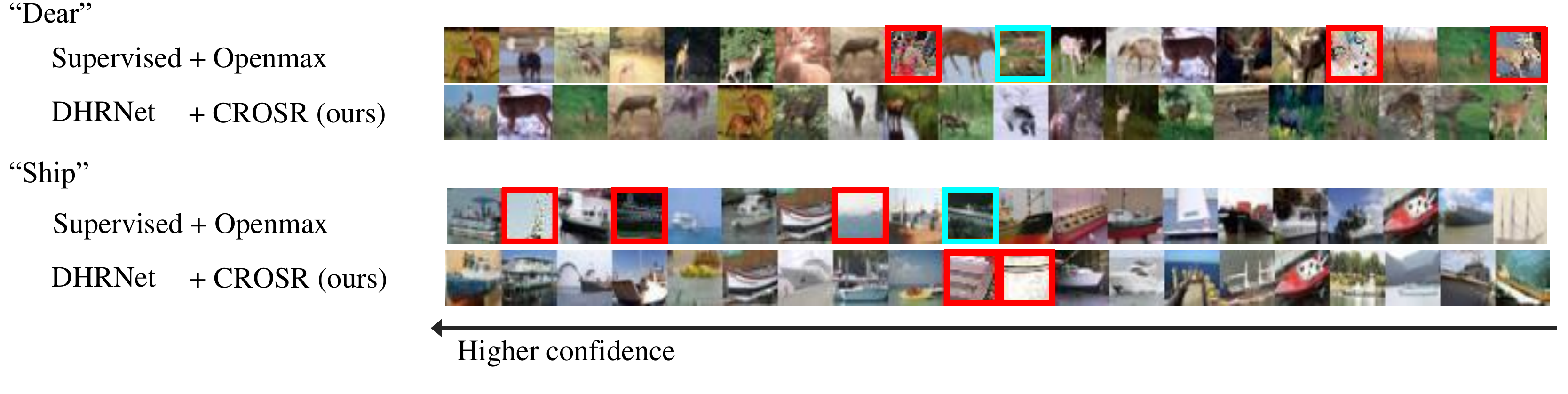}
  \end{center}
  \vspace{-5mm}
  \caption{\yoshia{Visualized samples. Sampled data points are sorted by each methods' confidence score, and the top samples are listed. \shaocvpr{The} red boxes show unknown samples, \yoshicvpr{and \shaocvpr{the cyan} ones show misclassification \shaocvpr{in} known classes}. \narm{Fewer unknowns to the left} indicate higher robustness.}}
  \vspace{-5mm}
  \label{fig:sorted}
\end{figure*}

\begin{table}[h]
	\caption{\yoshitxt{Open-set text classification results for DBpedia. F1-scores are shown \arma{for} various train/test class ratio\arma{s}.}}
   \footnotesize
  \begin{center}
  \begin{tabular}{|c||c|c|c|c|} \hline
  	   Method            & 4/14 & 4/12 & 4/8 & 4/4 \\ \hline \hline
       DOC               & 0.507 &  0.568 & {0.733} & 0.985  \\
       Softmax           & 0.460 & 0.503 & 0.662 & {\bf 0.988} \\
       Openmax           & 0.532 & 0.574 & 0.729 & 0.986 \\
       \sysname~(ours) & {\bf 0.582} & {\bf 0.627} & {\bf 0.765} & 0.987 \\
     \hline
  \end{tabular}\vspace{-8mm}
  \end{center}
  \label{tab:resdbpedia}
\end{table}

\vspace{1mm} \noindent {\bf MNIST}　\hspace{1mm}
MNIST is the most popular hand-written digit benchmark. It has 60,000 images for training and
10,000 for testing from \arma{ten} classes.
Although near-100\% accuracy has been achieved in closed-set classification~\cite{cirecsan2010deep}, 
\arm{the} open-set extension of MNIST remains a challenge
due to {the} variety of possible outliers.

As outliers, we used datasets of \arma{small gray-scale} images, namely Omniglot, Noise, and MNIST-Noise.
Omniglot is a dataset of hand-written characters from 
\arm{the} alphabets of various languages. 
We only used the test set because the outliers are only needed in the test phase. 
`Noise' is a set of images we synthesized by sampling 
each pixel value independently from a uniform distribution 
on [0, 1].
MNIST-Noise is also a synthesized set, \arm{made} by superimposing MNIST's
test images on Noise, and thus \arma{its images} are more similar to the inliers.
\yoshib{Figure~\ref{fig:samples} shows their samples.}
\arm{Each dataset} has 10,000 test images, the same as MNIST,
and this makes the known\arma{-to-}unknown ratio 1:1. 

We used a seven-layer plain CNN for MNIST. 
It consists of five convolutional layers with $3\times3$ kernels and 100 output channels, followed by ReLU non-linearities.
\shaocvpr{Max \shao{pooling} layers with \arma{a} stride of 2 are inserted after every two convolutional layers.} At the end of the convolutional layers, we put two fully connected layers with 
500 and 10 units, and the \arma{last one was} directly exposed to the Softmax classifier.
In \shaocvpr{DHRNet}, lateral connections are put after every pooling layer. The dimensionalit\arma{ies} of \arma{the} latent representations ${\bf z}_l$ \arma{were} all fixed to 32. 

\vspace{1mm} \noindent  {\bf CIFAR-10}　\hspace{1mm}
CIFAR-10 has 50,000 natural images for training and 10,000 for testing. \rei{\shaocvpr{It} consists of \arma{ten} classes, containing 5,000 training images for each class.} 
In \shaocvpr{CIFAR-10}, each class has large intra-class diversities by color, style, or pose difference, and state-of-the-art deep nets \arma{make a fair number of} classification errors within known classes.

We examined two types of network, a plain CNN and DenseNet~\cite{huang2017densely}, 
a state-of-the-art network for closed-set image classification.
The plain CNN is a VGGNet~\cite{simonyan2014very}-style network re-designed for CIFAR, and \arma{it} has 13 layers. 
\arm{The layers} are grouped \arm{into} three convolutional and one fully connected block.
The output channels of each convolutional block \arm{number}
64, 128, and 256, and they consist of two, two, 
and four convolutional layers with the same configuration.
All convolutional kernels are $3 \times 3$.
\arm{We set the depth of DenseNet to} 92 and \arma{the} growth rate \arm{to} 24.
The dimensionalit\arma{ies} of \arma{the} latent representations ${\bf z}_l$ \arma{were} all fixed to 32, the same as in MNIST. 

We used \arm{the} outliers collected by~\cite{liang2017enhancing} from other datasets,
\yoshib{\arma{i.e., }ImageNet and LSUN, and \arma{we} resized or cropped \arma{them} \arm{so that they would have the same sizes \footnote{URL: \url{https://github.com/facebookresearch/odin}.}}}
Among \arm{the outlier sets used} in~\cite{hendrycks2016baseline},
we did not use synthesized sets of Gaussian and Uniform because
they can be easily detected by baseline outlier-removal techniques.
\arm{The datasets each have} 10,000 test images, \shao{which} is the same as \arma{in} MNIST
and this makes the known\arma{-to-}unknown ratio 1:1. 

\vspace{1mm} \noindent  {\bf SVHN and TinyImageNet}　\hspace{1mm}
\yoshicvpr{
\shaocvpr{SVHN is a dataset of 10-class digit \yoshicvpr{photographs}, and TinyImageNet is a 200-class subset of ImageNet.}
In these datasets, we compare \sysname~\arma{with} recent GAN-based methods~\cite{ge2017generative,neal2018open} that utilize {\it unknown} training data synthesized by GANs.
A concern in the comparisons \arma{was the instability of the} training and resulting variance in the quality of \arma{the} training data \arma{generated by the GAN-based mechanisms}, which may make comparisons hard~\cite{lucic2017gans}.
Thus, we exactly follow\arma{ed} the evaluation protocols used in~\cite{neal2018open} (class separation within each single dataset, averaging over five trials, area-under-the-curve criteria), and directly compare\arma{d} our results against the reported numbers.
Our backbone network \arma{was} the same as the one used in~\cite{neal2018open} that consists of nine convolutional layers and one fully connected layers, except that ours \arma{had} decoding parts as shown in Eqn.~\ref{eqn:encdec}.}

\vspace{1mm} \noindent  {\bf DBpedia}　\hspace{1mm}
\yoshitxt{\arma{The} DBpedia ontology classification dataset contains 14 
classes of Wikipedia articles, 40,000 instances for training and 5,000 for testing.
We select\arma{ed} this dataset \arma{because it has the} largest number of classes
among \arma{the often-used} datasets in the literature of \arma{the} convnet-based large-scale text classification~\cite{zhang2015character} 
and \arma{for ease} in making various class splits. 
We conducted \arma{the} open-set evaluation with class separation using \shao{4} random classes as knowns 
and 
\shao{4, 8, and 10 as unknowns.}}

\yoshitxt{In DBpedia, we implemented \shaocvpr{DHRNet} on the basis \shao{of} a shallow-and-wide convnet~\cite{kim2014convolutional}, which \arma{had} three convolutional
layers with kernels whose sizes \arma{were} \shao{3, 4, and 5}, and whose output \shao{dimension} \arma{was} 100.
Text-classification convnets are extendable to \shaocvpr{DHRNet} by setting \hbox{$W = \textrm{(maximum text length)}$} and $H = 1$ in Fig.~\ref{fig:conv}.
The dimensionality of its bottleneck was 25.
We also implemented DOC~\cite{shu2017doc} using the same architecture as ours for \arma{a} fair comparison.}

\vspace{1mm} \noindent  {\bf Training \shaocvpr{DHRNet}}　\hspace{1mm}
We confirmed that \shaocvpr{DHRNet} can be trained by \arm{using}
the joint classification-reconstruction loss.
We used the SGD solver with learning-rate scheduling tuned in each dataset.
We set the weights of \arm{the} reconstruction loss and the classification loss to the same value 1.0.
\yoshicvpr{In principle, the weight of reconstruction error should be as large as possible while keeping the close-set validation accuracy, which would give the most regularized and well-fitted model. However, we obtained satisfactory results with the default value and did not tune them further.}
The \yoshicvpr{closed-set} test error\arma{s} of the networks \arm{for} each dataset \arma{are} \arm{listed} 
in Table~\ref{tab:closedacc}.
All of the networks were trained without \arm{any} large degradation in closed-set accuracy from the original ones. \arma{This and the subsequent} experiments were conducted using Chainer~\cite{tokui2015chainer}.

\vspace{1mm} \noindent  {\bf Weibull distribution fitting}　\hspace{1mm}
We used $\texttt{libmr}$ library \cite{Scheirer_2011_TPAMI} to compute \rei{the parameters} in Weibull distribution.
It has \arm{the} hyperparameters  $\alpha$ \arm{from} Eqn.~\ref{eqn:openmax} and
$\texttt{tail\_size}$, the number of extrema used to define the tails of the distributions.
We used the \arm{values} suggested in~\cite{bendale2016towards},
namely $\alpha = 10$ and $\texttt{tail\_size} = 20$.
\yoshi{For MNIST and CIFAR-10},
we did not use the rank calibration with $\alpha$ in Eqn.~\ref{eqn:weibull}, since it does not improve the performance \yoshi{due to \arm{the small} number of classes.}
For DenseNet in \shaocvpr{CIFAR-10}, we noticed that Openmax performed worse with the default parameters\arm{, so} we changed  $\texttt{tail\_size}$ to $50$.
Since heavily tuning these hyperparameters for specific types of outlier \arm{runs counter to} the motivation of 
open-set recognition \arm{for handling} \shao{\it unknowns},
we did not tune them for each \arm{of the} test sets. 

\vspace{1mm} \noindent  {\bf Results}　\hspace{1mm}
We show the results \arm{for} MNIST in Table~\ref{tab:resmnist}, 
for CIFAR-10 in Table~\ref{tab:resc10},
and for DBpedia in Table~\ref{tab:resdbpedia}.
The reported \arm{values} are F1-scores~\yoshi{\cite{sasaki2007truth}}
of known classes and {\it unknown} as a class with
a threshold 0.5.
\sysname~outperformed all of
the other methods consistently except \arma{in}
two settings.
\yoshia{Specifically, in MNIST, \sysname~outperformed Supervised + Openmax by more than 10\% in F1-score when using Omniglot or MNIST-noise as outliers, \arma{whereas it}
slightly underperformed with Noise, the easiest outliers. 
\arma{\sysname~}also performed better than or \arma{as well as the} stronger baselines LadderNet + Openmax and DHRNet + Openmax.
In CIFAR-10,}
the results \arm{for varying} thresholds are also shown in Fig. \ref{fig:f1_th}, in which \arm{it is clear that \sysname~outperformed the other methods} regardless of the threshold.

Interestingly, LadderNet with Openmax outperformed \arm{the} supervised-only networks. For instance,
LadderNet-Openmax achieved \arm{an 8.4\% gain in} F1-score in the MNIST-vs-Omniglot
setting and a 10.1\% gain in the MNIST-vs-MNIST-Noise setting.
This means regularization \arm{using the} reconstruction loss
is beneficial \arm{for} unknown detection; in other words,
using supervised losses in known classes is not the best for 
training open-set deep networks. 
However, \arm{no gains were had by adding only the reconstruction-error
term \arma{to} training objectives} 
in the natural image datasets. 
This means we need to use the reconstructive factors in the networks in a more explicit form by adopting \shaocvpr{DHRNet}.

\yoshitxt{For DBpedia, \sysname~outperformed the other methods, except when the number of train/test classes \arma{was} 4/4, \shao{which} is equivalent to \arma{the} closed-set settings. While DOC and Openmax performed almost on a par with each other, the improvement \arma{of} \sysname~\arma{over} Openmax was \arma{also} significant in this dataset.}

\vspace{-3mm}
\paragraph{\yoshicvpr{Comparison \arma{with} GAN-based methods}}
\yoshicvpr{
Table~\ref{tab:resgan} summarizes the results of ours and the GAN-based methods. 
Ours outperformed all of the other methods in MNIST and TinyImageNet, 
and all except Counterfactual in SVHN. 
While the relative improvements are within the ranges of the error bars, these results still means that our method, which does not use any synthesized training data, can perform on par or slightly better than the state-of-the-art GAN-based methods.}

\begin{table}[t]
	\vspace{-1mm} \caption{\yoshicvpr{Comparisons \arma{of} \sysname~\arma{with} recent GAN-based methods~\cite{ge2017generative}.}}
	\vspace{-2mm}
   \footnotesize
  \begin{center}
  \begin{tabular}{|c||c|c|c|} \hline
  	   Method / dataset      & MNIST & SVHN & TinyImageNet \\ \hline \hline
       Openmax               & 0.981 $\pm$ 0.005  &  0.894 $\pm$ 0.013 & {0.576}  \\
       G-Openmax             & 0.984 $\pm$ 0.005 &  0.896 $\pm$ 0.017 & {0.580}  \\
       Counterfactual        & 0.988 $\pm$ 0.004 &  {\bf 0.910} $\pm$ 0.010 & {0.586}  \\
       \sysname~(ours)     & {\bf 0.991 $\pm$ 0.004} & {0.899} $\pm$ 0.018 & {\bf 0.589}  \\
     \hline
  \end{tabular}\vspace{-8mm}
  \end{center}
  \label{tab:resgan}
\end{table}

\vspace{-4mm}\paragraph{\arm{In} combination with anomaly detectors}
To investigate how latent representations can be exploited more effectively, 
we replaced the $\ell^2$ distance in Eqn.~\ref{eqn:oneclass} by one-class learners.
We used the most popular one-class SVM (OCSVM) and Isolation Forest (IsoForest). For simplicity, we used the default hyperparameters
in $\texttt{scikit-learn}$~\cite{scikit-learn}. The results are shown in Table~\ref{tab:resocl}.
It \arm{reveals} that OCSVM had a more than 15\% gain \arma{in} F1-score in synthesized outliers, while it caused \arma{a} \arm{9\%} degradation in Omniglot.
Although we did not find an anomaly detector that \arm{consistently gave} performance improvements \arm{on all the} datasets, \arm{the results are still} encouraging.
\shao{The results suggest that} \shaocvpr{DHRNet} encodes more useful information
that \shao{is} not fully exploited by the per-class centroid based outlier
modeling.

\begin{table}[t]
    \vspace{-2mm}
	\caption{Open-set classification results for MNIST with different unknown detectors. \arma{Larger values are} better.}
  \footnotesize
  \vspace{-2mm}
  \begin{center}
  \begin{tabular}{|l||c|c|c|} \hline
  	  UNK detector & Omniglot & Noise & MNIST-noise \\ \hline \hline
      Supervised +            & &  &  \\
      \hspace{10mm} --$\ell^2$        & 0.680 & 0.890 & 0.720 \\
      \hspace{10mm} --OCSVM     & 0.647 & 0.899 & 0.919 \\ \hline
      Our DHRNet +            & &  &  \\
      \hspace{10mm} --$\ell^2$        & {\bf 0.793} & 0.826 & 0.827 \\
      \hspace{10mm} --OCSVM     & 0.702 & {\bf 0.979} & {\bf 0.976} \\
      \hspace{10mm} --IsoForest  & 0.649 & 0.908 & 0.839 \\
     \hline
  \end{tabular}\vspace{-7mm}
  \end{center}
  \label{tab:resocl}
\end{table}

\vspace{1mm} \noindent  {\bf Visualization}　\hspace{1mm}
Figure~\ref{fig:sorted} shows the test data from the known and unknown classes, 
sorted by the models' final confidences computed by Eqn.~\ref{eqn:openmax}.
In this figure, unknown data \arm{at} higher order mean that
the model is deceived by that data.
\yoshi{\arm{It is clear that our methods gave} lower confidences to the unknown samples, and \arm{they were deceived only by samples that had high similarity to the inlier.}}

\begin{figure*}[t]
  \begin{center}
    \hspace{-4mm}
    \includegraphics[width=520pt]{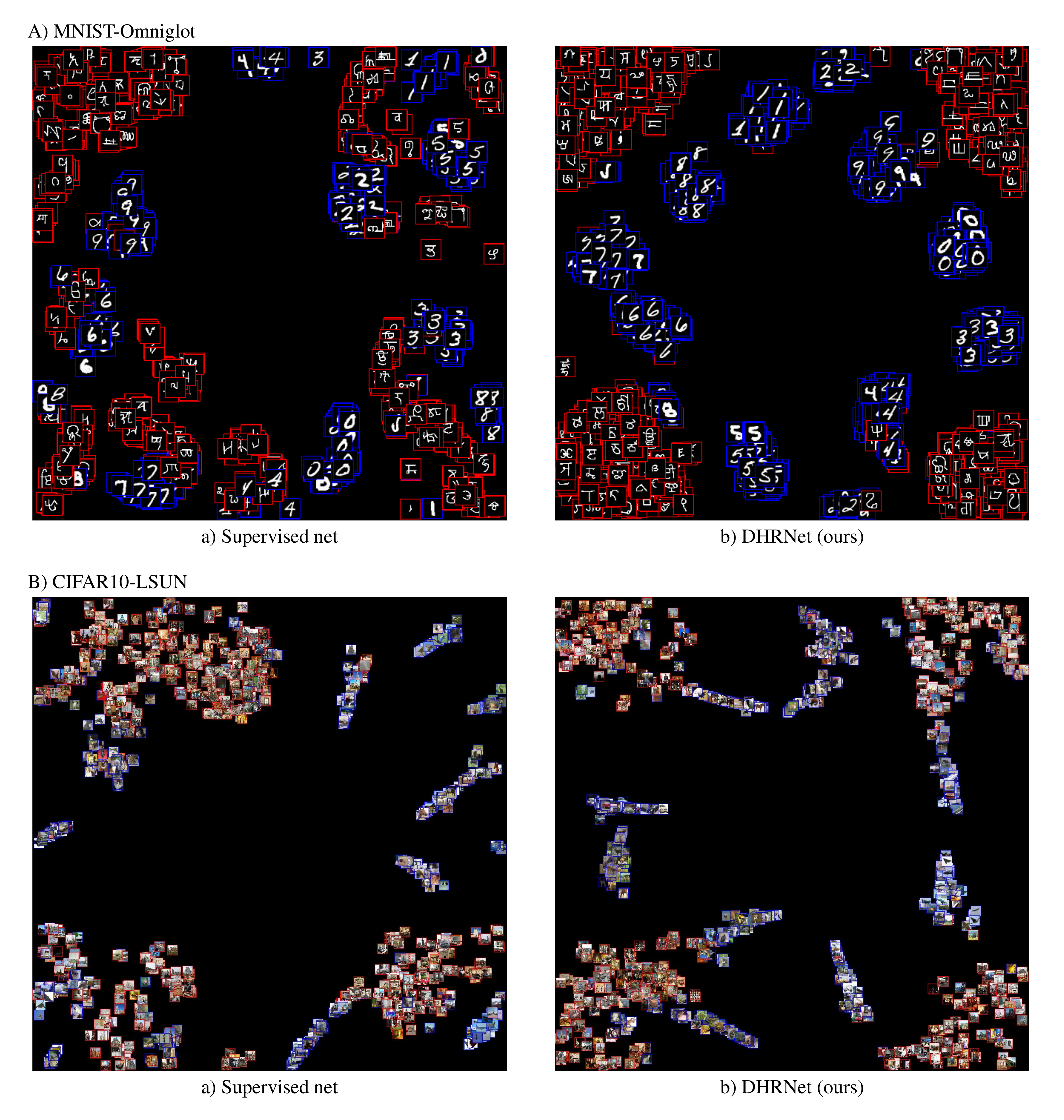}
  \end{center}
  \vspace{-3mm}
  \caption{Distributions of the known- and unknown-class images from the test sets over the representation spaces. Images with blue frames are known samples, and ones with red are unknowns. With the representations from our DHRNet, which contain both the prediction $\bm{y}$ and reconstruction latent variables $\bm{z}$, 
  \rei{the clusters of knowns and unknowns are more clearly separated.}
  }  
  \vspace{-5mm}
  \label{fig:embed}
\end{figure*}

We additionally visualize the learned representations by using t-distributed stochastic neighbor
embedding (t-SNE)~\cite{maaten_visualizing_2008}.
Figure~\ref{fig:embed} shows distributions of the representations
extracted from known- and unknown-class images in the test sets, embedded into two-dimensional planes.
\rei{Here we compare the distributions of the prediction $\bm{y}$} \rei{from the supervised net} and that of the concatenation of the prediction and the latent variable $[\bm{y}, \bm{z}]$ \rei{from our DHRNet. Their usages are shown in Eqns. (4) and (6) of the main text.}
While the existing deep open-set classifiers
exploit only $\bm{y}$, our CROSR exploits $[\bm{y}, \bm{z}]$. 
With the latent representation, 
\rei{the clusters of knowns and unknowns are more clearly separated,}
and \rei{this suggests that the representations learned by our DHRNet are  preferable for open-set classification}.

\vspace{1mm} \noindent  {\bf Run time}　\hspace{1mm}
\yoshicvpr{Despite of the extensions we made to the network, \sysname's
computational cost in \arma{the} test \arma{was} not much larger than Openmax's.
Figure~\ref{tab:time} shows the run times, which were computed on a single GTX Titan X graphic processor.
The overhead \arma{of} computing \arma{the} latent representations \arma{was} 
as small as 3--5 ms/image, negligible \arma{in relation} to the original cost when the backbone network is large.}
\begin{table}[t]
	\caption{\yoshicvpr{Run times of the models (milli seconds/image). The times were measured in CIFAR-10 with \arma{a} batch size $= 1$. 
	}}\vspace{-2mm}
  \footnotesize
  \begin{center}
  \begin{tabular}{|c|c|c|} \hline
   	Method / Architecture & Plain CNN & DenseNet \\ \hline \hline
     Softmax & 9.3 & 63.2 \\
     Openmax & 11.7 & 69.4  \\
      \sysname~(ours) & 16.5 & 72.4  \\ 
     \hline
  \end{tabular}
  \vspace{-8mm}
  \end{center}
  \label{tab:time}
\end{table}

\vspace{-2mm} \section{Conclusion} \vspace{-2mm}
We \arma{described} \sysname, a deep open-set classifier
augmented by latent representation learning for reconstruction.
To enhance \arm{the} usability of latent representations for unknown detection,
we also developed a novel deep hierarchical reconstruction net architecture.
Comprehensive experiments \arm{conducted on} multiple standard datasets demonstrated
that \arm{\sysname} outperforms previous state-of-the-art 
 open-set classifiers \arm{in most cases}.

\small

\vspace{-2mm} \section*{\normalsize Acknowledgement}\vspace{-2mm}
This work is in part
supported by JSPS KAKENHI Grant Number JP18K11348,
and Grant-in-Aid for JSPS Fellows JP16J04552. The authors
would like to thank Dr. Ari Hautasaari for his helpful
advice to improve the manuscript.

{\small
\vspace{-2mm} \bibliographystyle{ieee}
\bibliography{mybib}
}

\end{document}